\documentclass[lettersize,journal]{IEEEtran}
\usepackage{amsmath,amsfonts}
\usepackage{algorithmic}
\usepackage{algorithm}
\usepackage{array}
\usepackage[caption=false,font=normalsize,labelfont=sf,textfont=sf]{subfig}
\usepackage{textcomp}
\usepackage{stfloats}
\usepackage{url}
\usepackage{verbatim}
\usepackage{graphicx}
\usepackage{cite}
\usepackage{orcidlink}
\usepackage{booktabs}
\usepackage{multirow}
\usepackage{pifont}
\newcommand{\cmark}{\ding{51}}%
\newcommand{\xmark}{\ding{55}}%
\begin{document}

\title{Assessing the Efficacy of Deep Learning Approaches for Facial Expression Recognition in Individuals with Intellectual Disabilities}

\author{F. Xavier Gaya-Morey \orcidlink{0000-0003-1231-7235}, Silvia Ramis \orcidlink{0000-0002-1039-4387}, Jose M. Buades-Rubio \orcidlink{0000-0002-6137-9558}, Cristina Manresa-Yee \orcidlink{0000-0002-8482-7552}
\thanks{F. Xavier Gaya-Morey, Silvia Ramis, Jose M. Buades-Rubio and Cristina Manresa-Yee are with the Computer Graphics and Vision and AI Group (UGIVIA) and the Research Institute of Health Sciences (IUNICS), from the Universitat de les Illes Balears, Carretera de Valldemossa, km 7.5, Palma, 07122, Illes Balears, Spain.

\textit{\textbf{This work has been submitted to the IEEE for possible publication. Copyright may be transferred without notice, after which this version may no longer be accessible.}}}}%



\maketitle

\begin{abstract}
Facial expression recognition has gained significance as a means of imparting social robots with the capacity to discern the emotional states of users. The use of social robotics includes a variety of settings, including homes, nursing homes or daycare centers, serving to a wide range of users. Remarkable performance has been achieved by deep learning approaches, however, its direct use for recognizing facial expressions in individuals with intellectual disabilities has not been yet studied in the literature, to the best of our knowledge. To address this objective, we train a set of 12 convolutional neural networks in different approaches, including an ensemble of datasets without individuals with intellectual disabilities and a dataset featuring such individuals. Our examination of the outcomes, both the performance and the important image regions for the models, reveals significant distinctions in facial expressions between individuals with and without intellectual disabilities, as well as among individuals with intellectual disabilities. Remarkably, our findings show the need of facial expression recognition within this population through tailored user-specific training methodologies, which enable the models to effectively address the unique expressions of each user.
\end{abstract}


\section{Introduction}

    Understanding the emotional state of people is important both on an individual and societal level. This understanding allows an effective communication, promotes empathy, or enhances social dynamics, among others. 
    Non-verbal communication through body language is a powerful form of communication that offers insights into the individual's emotional state \cite{degelder}. Within the body language, facial expressions stand out as a particularly significant component and although facial expressions are not emotions themselves \cite{barrett2019emotional}, they provide a compelling visual representation. 
    Providing artificial cognitive systems, such as social robots, with the ability to discern emotions can enhance the user experience in social contexts \cite{dinuovo, limaTrans}. Research in affective computing has notably focused on automated facial expression recognition (FER), that is, the identification and analysis of human facial expressions \cite{ko2018brief, revina2021survey, li2022deep}. A significant body of research builds upon  Ekman's six fundamental and universally recognized facial expressions: anger, happiness, surprise, disgust, sadness, and fear \cite{ekmanuniversal}, although their universality has been a topic of debate \cite{Barrett} and they do not express the wide range of expressions that humans can do. Examples of applications of automated FER extend across an array of domains, including medical diagnosis and treatment \cite{grabowski2019emotional}, as well as its integration into the fields of Human-Computer Interaction (HCI) and Human-Robot Interaction (HRI) \cite{medjden2020adaptive, ramis2020using}.
  
    Deep learning (DL), particularly Convolutional Neural Networks (CNNs), has recently emerged in the FER domain \cite{li2022deep}. These DL models have demonstrated remarkable performance, often attaining state-of-the-art results in FER tasks. This transition to DL signifies a paradigm shift from early FER research, which focused on  facial feature extraction and the analysis of facial appearance \cite{pantic2000automatic, fasel2003automatic}. The adoption of DL techniques has achieved important advancements in FER; however, these techniques often require large amounts of high-quality labeled data for training to generalize well to new, unseen cases.

    In the case of people with intellectual disability, that is, people that present limitations in their ability to learn at an expected level and function in daily life, the identification of facial expressions is an important challenge \cite{ murray2019}, especially for  those that present verbal communication limitations \cite{ADAMS2011293}. Understanding how this population feels is of utmost importance as it can indicate discomfort, sadness or pain. The severity of their limitations can vary widely \cite{WorldHealthOrganizationWHO2018}, and the literature shows that  their facial expressions may differ from those without intellectual disabilities \cite{Zaja2008, Rayworth2997, Wilczenski}. The direct application of existing FER systems for this population would speed up the design of behavioral and cognitive capabilities in artificial agents at homes, nursing homes, day care centers or hospitals. However, a notable gap persists in the literature regarding comprehensive studies assessing the applicability of these systems to people with intellectual disabilities. Further, there is a lack of tailored solutions specifically addressing the needs of this population, likely due to the scarcity of high-quality datasets.  

     In this work, we articulate three central research questions:
    
    \begin{enumerate}
        \item Can DL models, originally trained with standard datasets of facial expressions, yield proficient performance when applied to individuals with intellectual disabilities? (Q1)
        \item Can DL models, specifically trained on a dataset comprising individuals with intellectual disabilities, accurately predict the facial expressions of other individuals with intellectual disabilities? (Q2)
        \item What disparities and commonalities exist in the facial expressions of individuals with and without intellectual disabilities, learned by DL models? (Q3)
    \end{enumerate}

    Answering these questions can provide substantial benefits across several domains, including inclusivity or adaptability. By assessing whether DL models trained on standard datasets perform adequately when applied to individuals with intellectual disabilities, we can identify any gaps in inclusivity and establish a performance baseline for existing technologies. This helps in understanding whether current AI systems can be deployed more broadly without extensive customization, enhancing their accessibility and cost-effectiveness. Secondly, evaluating models specifically trained on data from individuals with intellectual disabilities can lead to increased accuracy and customization in AI interactions, ensuring the technology is sensitive to the group characteristics. Lastly, exploring the differences and commonalities in facial expressions between individuals with and without intellectual disabilities enriches our understanding of AI's ability to interpret human emotions universally. This knowledge is vital for improving dataset compilation or designing fairer AI systems. 
    
    The work is organized as follows: Section II presents a comprehensive review of related works, highlighting the lack of FER-related works using DL techniques for people with intellectual disabilities. Section III describes the methodological approach, encompassing dataset curation, model selection, data preprocessing, an overview of our XAI strategy, and a description of the experiments conducted. We then present in Section IV and V our empirical findings and results, and present a comprehensive discussion of these outcomes. Finally, we summarize the main findings and present future work lines.

\section{Related work}

Recent years have witnessed remarkable developments in FER, with a shift towards the application of  DL techniques, particularly CNNs \cite{mellouk2020facial}. Current research works show the efficacy of these systems achieving high performance recognizing from  6 to 8 facial expressions over well-known datasets such as CK+ or JAFFE \cite{li2022deep}. These datasets, comprised by actor posed expressions or labelled images, do not specifically inform about the disabilities.

Considering people with moderate to severe intellectual impairments, who can present differences when producing facial expressions, we find a scarce number of works that aim at recognizing automatically their basic facial expressions, especially with DL techniques. 

Working in a similar line addressed to people with profound intellectual and multiple disabilities (PIMD), Campomanes-Álvarez and Campomanes-Álvarez \cite{Campomanes} designed a recognizer based on analyzing the changes in the appearance of three face regions -eyes, mouth and jaw- using different techniques on each zone (e.g.  random forest, k-nearest neighbour, naïve bayes, logistic regression, neural network). They built four datasets, labeled with the regions’ appearance, to test each recognizer: one with 375 samples for the eyes and the eyebrows which included 6 people with PIMD and 5 other people, one for the mouth with 375 samples which included 6 people and examples from CK+; and one for the jaw region with 1378 samples, that included 5 people. Labels for each zone indicated appearance comprising different states for the eyes (closed, semi, widened, winking or neutral), the eyebrows (frown, raising or neutral), the mouth (corners of mouth up, corner of mouth down, mouth wide open, lips movements and neutral state) and the jaw (grinding, biting, drooping  and no-movement).  Their results were promising, however, the evaluation did not include the basic facial expressions identifications and the datasets were lacking in the inclusion of individuals with PIMD.

Dovgan et al. \cite{dovgan} developed a decision support system (DSS)  to recognize behaviour patterns, including facial expressions, of two persons with PIMD. They used 8 machine learning techniques (e.g. random forest, support vector machine, K-nearest neighbours) and a set of rules informed by experts (e.g. caregivers). They fused the techniques using an ensemble approach, achieving 12 ensembles. Their results were modest when classifying the inner state (neutral, pleasure or displeasure)  of the individuals (max. accuracy of 62.4 for one case and 69.3 for the other one). They commented limitations such as difficulties in the label annotations (even for caregivers), unbalanced data, facial expressions not well recognized by the system due to the conditions of the recordings (the experiment took place during normal care activities), informative behaviours for the caregivers that are not always recognized by the system or inconsistency in the person’s behaviour.

Addressing other populations which present intellectual impairments, whose facial expressions may be similar to the ones of people without disabilities \cite{Smith1996}, Paredes et al. \cite{ Paredes_2023}  built a system  based on CNNs  aiming at  identifying anger, happiness, sadness, surprise, and neutrality in people with Down Syndrome. They compiled a dataset including 1200 images of 8- to 12-year-old Down syndrome individuals displaying spontaneous emotions with their therapist or tutor during daily activities and achieved an average accuracy of 91.4\%, with happiness being the best recognized expression.

Despite advances in FER, the research literature reveals a shortage of studies focused on the direct application of FER models to diverse populations, such as individuals with intellectual disabilities.

\section{Methods}

    This section describes the datasets, image processing procedures, DL models, and XAI techniques employed in this study. Additionally, it outlines the various experiments conducted to address the research questions.

    \subsection{Datasets}
    \label{sec:datasets}

        Seven datasets were employed in this study, which can be found in Table \ref{table:datasets}. The initial four datasets are well-established benchmarks commonly used in FER research, namely the Extended Cohn-Kanade (CK+) \cite{lucey2010extended}, BU-4DFE \cite{yin20063d}, JAFFE \cite{lyons1998japanese}, and WSEFEP \cite{olszanowski2014warsaw} datasets. The CK+ dataset comprises 593 sequences collected from 123 subjects, with each sequence labeled with one out of seven facial expressions: anger, contempt, disgust, fear, happiness, sadness, and surprise. On the other hand, the BU-4DFE dataset includes 606 sequences from 101 subjects, each subject contributing six sequences, one for each facial expression (anger, disgust, fear, happiness, sadness, and surprise). The JAFFE and WSEFEP datasets also offer a neutral class, in addition to these six facial expressions. The first one encompasses 213 images taken from 10 female Japanese actresses, while the second one comprises 210 images from 30 individuals.

        Two additional datasets were included in the study: FEGA and FE-test \cite{SilNet2022}. The FEGA dataset is noteworthy for its multi-label annotations, including facial expression, gender, and age. This dataset involves 51 subjects, each performing the same seven facial expressions as JAFFE and WSEFEP, with eight repetitions for each expression, resulting in multiple snapshots. Conversely, the FE-test dataset consists of 210 frontal images captured "in the wild". For the experiments, we merged the BU-4DFE, JAFFE, WSEFEPF, CK+ and FEGA datasets, this union will be referred to with the name of "FER-DB5", for simplicity.
        
        Lastly, the study also made use of the MuDERI dataset \cite{shukla2016muderi}. The MuDERI dataset is a comprehensive multimodal database featuring 12 participants with intellectual disabilities. This dataset encompasses two audio-visual recordings for each participant. In the first recording, participants were exposed to positive stimuli to elicit positive emotions, while the second recording involved negative stimuli to evoke negative emotions. These videos are segmented by timestamps, and each timestamp is annotated with three facial expressions: happiness, sadness, and anger. Additionally, the dataset includes annotations of electroencephalography (EEG) signals, electrodermal activity (EDA) signals, and Kinect data that were synchronized with the audio-visual recordings using these timestamps. 

        \begin{table*}
        \centering
        \caption{Detailed list of datasets used in the study.}
        \label{table:datasets}
        \begin{tabular}{llll|lll|llllllll|ll}
            \multicolumn{4}{c|}{\textbf{Details}} & \multicolumn{3}{c|}{\textbf{Data}} & \multicolumn{8}{c|}{\textbf{Classes}} & 
            \multicolumn{2}{c}{\textbf{Used for}}\\
            \multicolumn{1}{c}{\rotatebox{90}{\textbf{Name}}} & \multicolumn{1}{c}{\rotatebox{90}{\textbf{Ref.}}} & \multicolumn{1}{c}{\rotatebox{90}{\textbf{Year}}} & \multicolumn{1}{c|}{\rotatebox{90}{\textbf{Authors}}} & \multicolumn{1}{c}{\rotatebox{90}{\textbf{Users}}} & \multicolumn{1}{c}{\rotatebox{90}{\textbf{Samples}}} & \multicolumn{1}{c|}{\rotatebox{90}{\textbf{Data type}}} &\rotatebox{90}{\textbf{Happiness}} & \rotatebox{90}{\textbf{Sadness}} & \rotatebox{90}{\textbf{Anger}} & \rotatebox{90}{\textbf{Disgust}} & \rotatebox{90}{\textbf{Fear}} & \rotatebox{90}{\textbf{Surprise}} & \rotatebox{90}{\textbf{Neutral}} & \rotatebox{90}{\textbf{Contempt}} &
            \rotatebox{90}{\textbf{Training}} & 
            \rotatebox{90}{\textbf{Testing}}\\ 
            \midrule
            FEGA & \cite{SilNet2022} & 2022 & Ramis et al. & 51 & 1668 & Image & \cmark & \cmark & \cmark & \cmark & \cmark & \cmark & \cmark & \xmark & \cmark & \xmark\\
            FE-Test & \cite{SilNet2022} & 2022 & Ramis et al. & 210 & 210 & Image & \cmark & \cmark & \cmark & \cmark & \cmark & \cmark & \cmark & \xmark & \xmark & \cmark\\
            MuDERI & \cite{shukla2016muderi} & 2016 & Shukla et al. & 12 & 24 & Video & \cmark & \cmark & \cmark & \xmark & \xmark & \xmark & \xmark & \xmark & \cmark & \cmark\\
            WSEFEP & \cite{olszanowski2014warsaw} & 2014 & Olszanowski et al. & 30 & 210 & Image & \cmark & \cmark & \cmark & \cmark & \cmark & \cmark & \cmark & \xmark & \cmark & \xmark\\
            CK+ & \cite{lucey2010extended} & 2010 & Lucey et al. & 123 & 593 & Video & \cmark & \cmark & \cmark & \cmark & \cmark & \cmark & \xmark & \cmark & \cmark & \xmark\\
            BU-4DFE & \cite{yin20063d} & 2006 & Yin et al. & 101 & 606 & Video & \cmark & \cmark & \cmark & \cmark & \cmark & \cmark & \xmark & \xmark & \cmark & \xmark\\
            JAFFE & \cite{lyons1998japanese} & 1998 & Lyons et al. & 10 & 213 & Image & \cmark & \cmark & \cmark & \cmark & \cmark & \cmark & \cmark & \xmark & \cmark & \xmark\\
            \bottomrule
        \end{tabular}
        \end{table*}

    \subsection{Image preprocessing}
    \label{sec:preprocessing}

        To facilitate the FER task to the models, we applied preprocessing to the images, comprising face detection, alignment, and cropping. The face detection is performed using the "a contrario" framework, a method proposed by Lisani et al. \cite{lisani2017contrario}. Then, we employed the 68 facial landmarks, as outlined by Sagonas et al. \cite{sagonas2013300}, to precisely locate the positions of the eyes and achieve facial alignment. This process involved calculating the geometric centroids of each eye to compute the necessary rotation angle for horizontal alignment of the eyes. After the facial alignment, we performed cropping to isolate the facial region of interest and then resized the images to match the specific dimensions required for compatibility with the CNNs employed in this research.

    \subsection{Models}
    \label{sec:models}

        A set of twelve neural network models was constructed for the classification of three basic facial expressions. These models encompass both widely recognized architectures and CNNs tailored explicitly for Facial Expression Recognition (FER). The nine established models are AlexNet \cite{Alexnet2012}, VGG16 and VGG19 \cite{VGG2015}, ResNet50 and ResNet101V2 \cite{ResNet2015}, InceptionV3 \cite{Inception2015}, Xception \cite{Xception2017}, MobileNetV3 \cite{MobileNetV3}, and EfficientNetV2 \cite{EfficientNetV2}. The three specialized FER models are: SilNet \cite{SilNet2022}, SongNet \cite{SongNet2014}, and WeiNet \cite{WeiNet2015}, denoted as such for convenience.

        All of these CNNs underwent training and evaluation using the datasets detailed in subsection \ref{sec:datasets}, following the preprocessing steps outlined in subsection \ref{sec:preprocessing}. The  selection of models presents a variety of architectures whose exploration and performance assessment could offer valuable insights to the research.

        All code was written in Python, and the Keras library was used to implement the AlexNet, WeiNet, SongNet and SilNet models layer by layer. The remaining models were directly accessible using the Keras API, along with their pre-trained weights on ImageNet.

    \subsection{XAI approach}

        To obtain further insights into the inner workings of the models when performing the FER task, we employed the XAI technique introduced in \cite{ijimai}, which obtains global per-class explanations by normalizing LIME explanations \cite{ribeiro2016why}.

        LIME stands for Local Interpretable Model-agnostic Explanations, and consists on the perturbation of different regions at image level, to infer the relevance of each region for the outputted prediction. We used SLIC \cite{achanta2012slic} to segment the images and obtain the regions (approximately 30), dark color to occlude the regions, and 1,000 samples for each explanation. For the sake of simplicity, we focused on the positive relevance, omitting the negative one.

        Then, following the procedure described in \cite{ijimai}, we fit all explanations onto a normalized space, where all face landmarks are located at the same coordinates. This allows for the aggregation of multiple explanations into different heat maps, by classes and by networks, to better understand the key regions used by the models in each case.

    \subsection{Experiments}

        In this section, we provide a comprehensive overview of the three experiments conducted to address the research questions posed in this study.

        \subsubsection{First experiment: Training on FER datasets}

            In the first experiment, we extended upon the methodology introduced in \cite{interaccion2023} across the entire set of networks detailed in Section \ref{sec:models}. The objective was to answer Q1, evaluating whether diverse networks, trained on extended datasets designed for the FER task, could accurately classify facial expressions in individuals with intellectual disabilities from the MuDERI dataset.

            This experiment included $k$ trainings with $k=15$ on FER-DB5, which combines five FER datasets (CK+, BU-4DFE, JAFFE, WSEFEP, and FEGA, as detailed in Section \ref{sec:datasets}). Each training was subjected to two evaluations: one on Google FE-Test and another on MuDERI. This evaluation aimed to validate the models on dissimilar datasets —one involving users with intellectual disabilities and the other without them. Additionally, $k$ trainings, also with $k=15$, were performed on the full MuDERI dataset, followed by evaluations on Google FE-Test. This second set of trainings allowed to investigate in more depth the inherent differences between the Google FE-Test and MuDERI datasets. The value for $k$ was set sufficiently big to enhance the robustness of the performance assessment and mitigate the impact of individual training variations.

        \subsubsection{Second experiment: training on MuDERI}
        \label{sec:experiment2}

\begin{figure}[h]
            \centering
            \includegraphics[width=\linewidth]{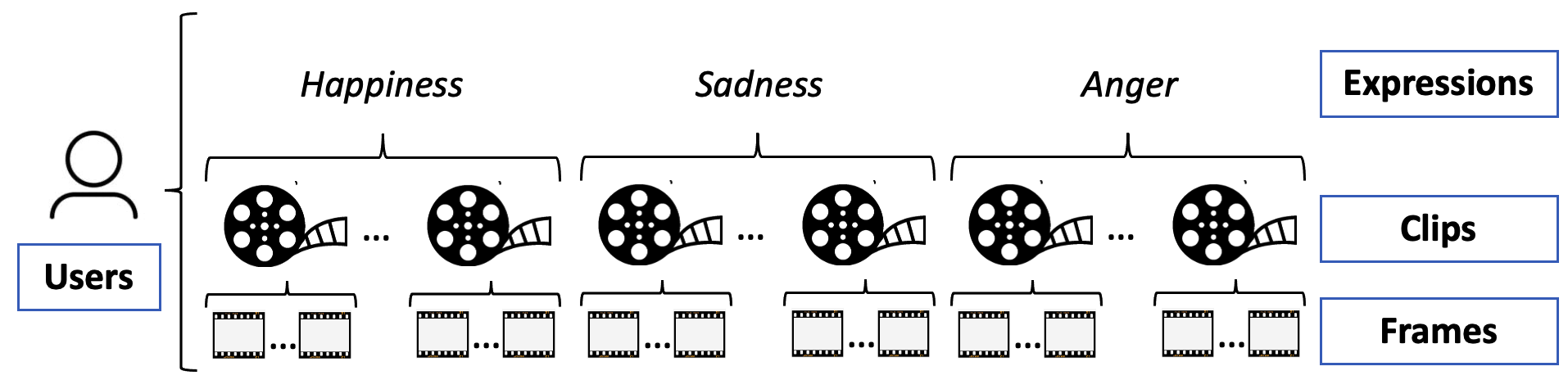}
            \caption{Considerations for the scenarios: each user has video clips for each expression (happiness, sadness and anger). The clips are then divided into frames.}
            \label{fig:esquema}
        \end{figure}

            The second experiment targeted the research question Q2 and involved training and evaluating models directly on the MuDERI dataset. Four distinct scenarios were explored based on the split between training and test sets considering users, clips and frames (see Figure \ref{fig:esquema}). This granular decomposition of the experiment aimed at controlling the limitations of the MuDERI dataset and explore various possibilities when dealing with individuals with intellectual disabilities. The scenarios were the following:

            \begin{enumerate}
                \item \textbf{User-based Split}: Training was conducted on a partition of MuDERI where the split was performed by users, training on some users and evaluating on the remaining ones.
                
                \item \textbf{Clip-based Split (Full Exposure)}: The split was done by clips, ensuring that models were exposed to all users in the training, thereby providing insights to user-specific facial expressions. The clips were randomly selected for the training or testing set. 
                
                \item \textbf{Clip-based Split (Restricted Exposure)}: The split was done by clips, ensuring that models were exposed to all users in the training. However, in this case, the training set included all users and all classes. This aimed at assessing the model's ability to recognize expressions for a user only if it has encountered other clips of the same user and class during training. Exceptionally, in the case of users with only a single clip for a specific class, this clip  was only included in the training set, not the test set. 
                
                \item \textbf{Frame-based Split}: The split was performed by frames, allowing adjacent frames to randomly fall into the training or testing sets.
            \end{enumerate}

        \subsubsection{Third experiment: explaining results}


            In the final experiment, we explored the differences in regions used by the models to identify facial expressions. We accomplished this by comparing the resulting heat maps by network and class,  from two distinct trainings: one on FER-DB5 and the other on MuDERI (clip-based split with restricted exposure). This comparison aimed at understanding the similarities and dissimilarities in FER for individuals with and without intellectual disabilities. Furthermore, for the trainings on FER-DB5, we presented heat maps computed by testing the models on Google FE-Test and MuDERI. This enables the observation of differences in important regions when the same models are employed for users with and without intellectual disabilities.

\section{Results}

    In this section, we present the results achieved in the experiments described in the previous section.

    \subsection{First experiment: training on FER datasets}

\begin{figure}[h]
            \centering
            \includegraphics[width=\linewidth]{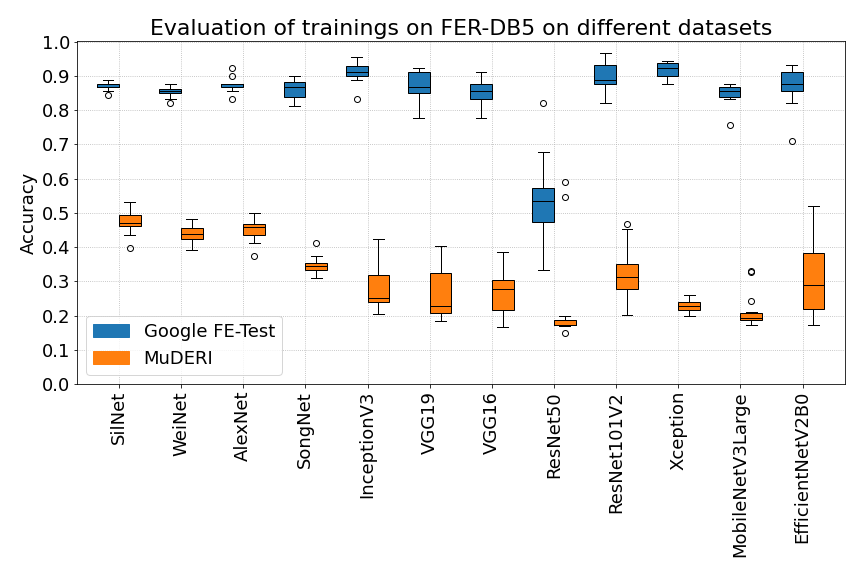}
            \caption{Box-plot showing the accuracy by network and evaluation dataset for the $k=15$ trainings on FER-DB5. The big gap in accuracy between evaluating the models on Google FE-Test and on MuDERI can be appreciated.}
            \label{fig:bd5-boxplot}
        \end{figure}

        Figure \ref{fig:bd5-boxplot} depicts the results of the evaluation of the trainings on the FER-DB5 dataset on Google FE-Test and MuDERI. The box plot illustrates the spread of results through different trainings, showcasing the median, quartiles, and outliers for each network.

        Based on the results, we present the following findings:
        \begin{itemize}
            \item None of the networks achieved satisfactory results on the MuDERI dataset, with accuracies consistently below 55\%.
            \item Almost all networks, excluding ResNet50, performed well on the Google FE-Test, with accuracies exceeding 80\% in most cases.
            \item ResNet50 exhibited inconsistent performance across trainings, with one achieving accuracy above 80\% on Google FE-Test.
            \item Trainings on MuDERI displayed larger box and whiskers compared to FER-DB5 for eight out of twelve networks, indicating higher variation in accuracy between iterations.
        \end{itemize}

\begin{figure}[h]
            \centering
            \includegraphics[width=\linewidth]{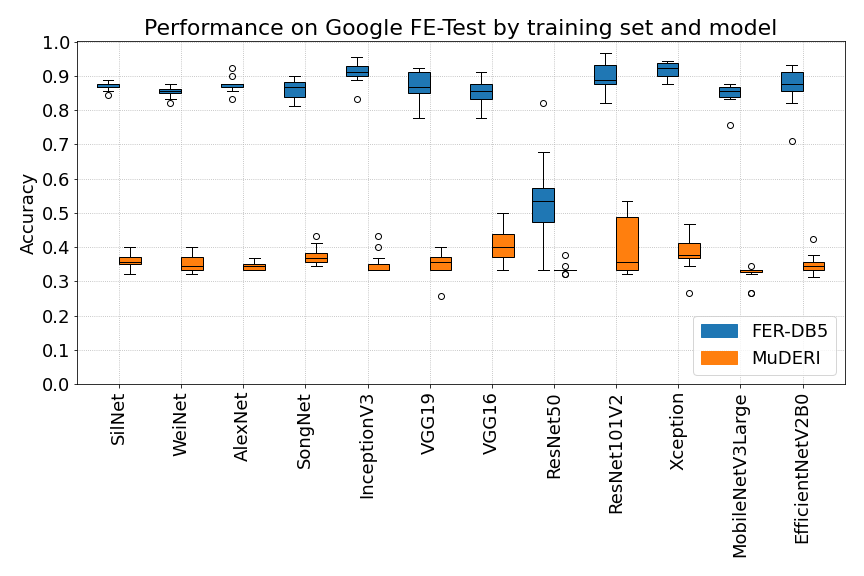}
            \caption{Accuracy on the Google FE-Test dataset of the $k$ (with $k=15$) trainings performed on FER-DB5 and MuDERI, by network. In this case, the low performance on Google FE-Test of the models trained on MuDERI should be noted.}
            \label{fig:google-boxplot}
        \end{figure}

        In Figure \ref{fig:google-boxplot}, a box plot illustrates the accuracy of $k=15$ trainings on FER-DB5 and MuDERI, evaluated on Google FE-Test. Results indicate poor performance for trainings on MuDERI, with accuracies mostly below 40\%.
        
\begin{figure}[h]
            \centering
            \includegraphics[width=\linewidth]{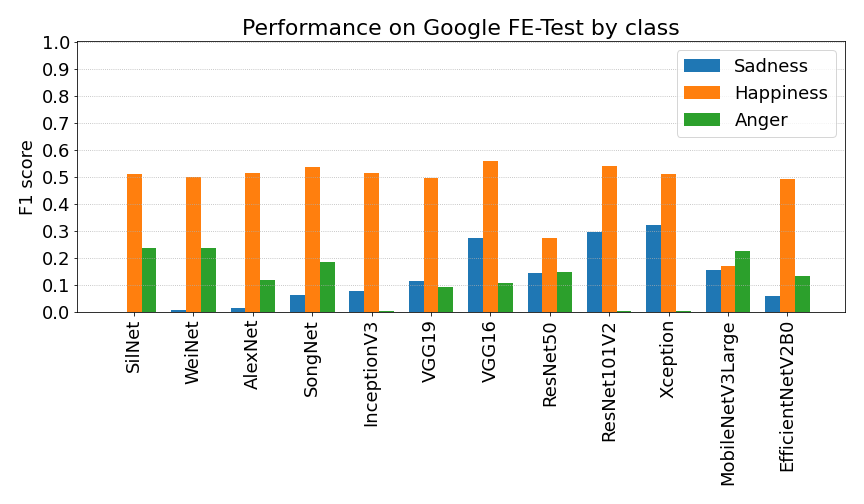}
            \caption{Average per-class F1 score obtained on Google FE-Test by the trainings on MuDERI. Note the great difference among classes.}
            \label{fig:google-per-class}
        \end{figure}

        To explore the results on MuDERI further, Figure \ref{fig:google-per-class} displays the average per-class F1 score on Google FE-Test. Significantly, the "happiness" class dominates detections, leading to low recall and F1 scores for other classes.

    \subsection{Second experiment: training on MuDERI}
        \label{sec:second_exp}
        
\begin{figure*}[h]
            \centering
            \includegraphics[width=0.7\textwidth]{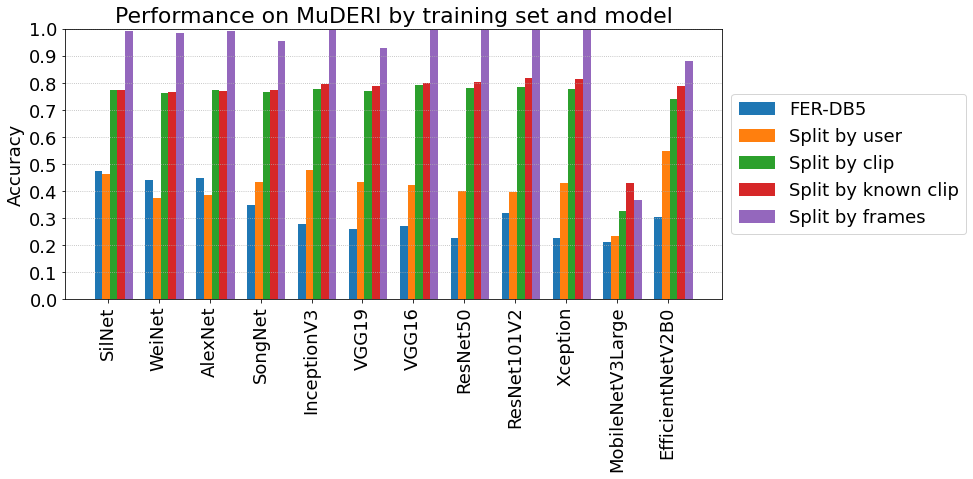}
            \caption{Accuracy of the different training scenarios on MuDERI, by network. We have added the average results obtained by the 15 trainings on FER-DB5 for comparison.}
            \label{fig:muderi-acc}
        \end{figure*}
\begin{figure*}[h]
            \centering
            \includegraphics[width=0.7\textwidth]{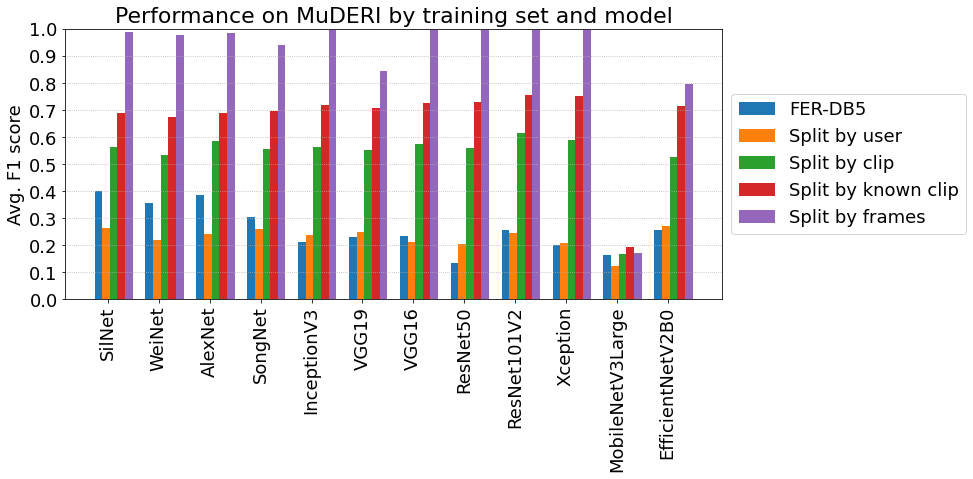}
            \caption{F1 score of the different training scenarios on MuDERI, by network. We have added the average results obtained by the 15 trainings on FER-DB5 for comparison.}
            \label{fig:muderi-f1s}
        \end{figure*}

        Figures \ref{fig:muderi-acc} and \ref{fig:muderi-f1s} depict the accuracy and F1 score, respectively, of different training scenarios on MuDERI, alongside results obtained by FER-DB5 trainings on average.

        The figures provide the following insights:

        \begin{itemize}
            \item The 1st training scenario (user-based split) yields the worst results, close to those achieved by FER-DB5 trainings.
            \item The 2nd and 3rd scenarios (clip-based split) show similar accuracy, with the 3rd scenario performing over 10\% better on the F1 score for all networks.
            \item The 4th scenario (frame-based split) attains the highest accuracy, but potential overfitting is suspected due to the similarity of consecutive frames.
            \item Training scenarios exhibit consistent results across different networks, except for MobileNetV3.
            \item EfficientNetV2 excels in the 1st scenario but underperforms in the 4th. MobileNetV3 consistently records the worst results.
        \end{itemize}

         These findings offer insights into the varying effectiveness of different training scenarios on MuDERI and highlight small network-specific performance variations.

    \subsection{Third experiment: explaining results}
        \label{sec:third_exp}

        In Figure \ref{fig:heatmaps} we display heat maps for all classes and networks, computed for the training on FEER-DB5 tested on Google FE-Test and MuDERI, and for the training and test on MuDERI. The following aspects can be observed:

        \begin{figure*}[h]
            \centering
            \includegraphics[width=.6\textwidth]{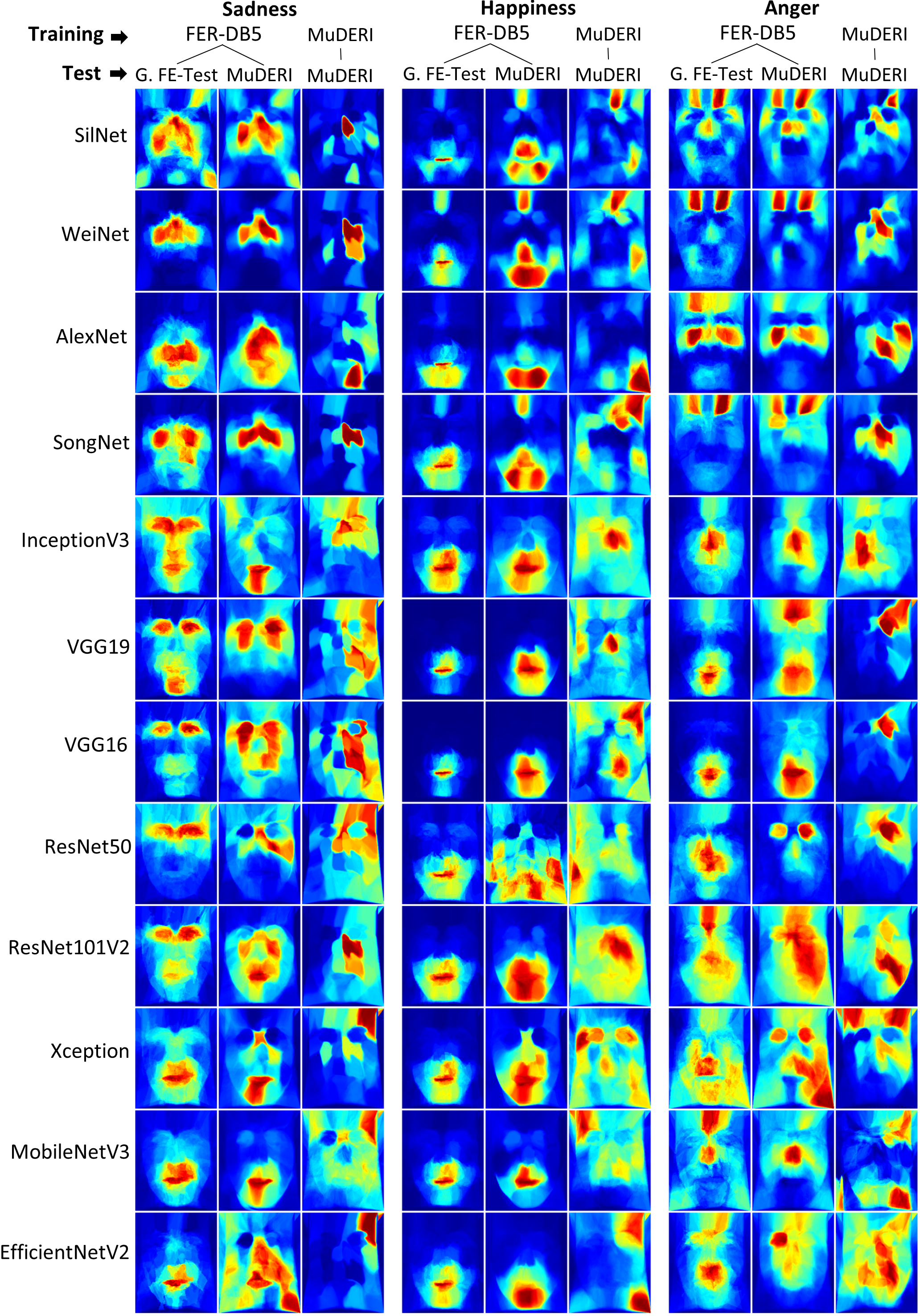}
            \caption{Computed heat maps for the different trainings on FEER-DB5 tested on Google FE-Test and MuDERI, and for the training and test on MuDERI, grouped by model (by rows), by classes (three major columns), and by training-test sets used (four sub-columns, for each expression). Warm tones like red and orange represent the important regions, while cooler tones like green and blue represent less important ones.}
            \label{fig:heatmaps}
        \end{figure*}
        
        \begin{itemize}

            \item \textbf{Training Set Influence}: The choice of the training set significantly impacts the resulting heat maps, leading to notable distinctions between trainings on FER-DB5 and MuDERI. Heat maps generated on MuDERI appear less intuitive and, in some instances, exhibit a degree of complexity.

            \item \textbf{Test Set Influence}: There are many similarities between heat maps for a specific network, class, and training set across various test sets.

            \item \textbf{Network Influence}: Considerable variation exists in the learned features across models, particularly notable for "Sadness" and "Anger" expressions and for the four networks without pre-trained weights (SilNet, WeiNet, AlexNet, and SongNet). However, some common patterns emerge, contingent on the training set, with clearer trends observed for the "Happiness" class trained on FER-DB5.
            
            \item \textbf{Class Influence}: Differences between classes are more pronounced in models trained on the MuDERI dataset. Conversely, models trained on FER-DB5 tend to emphasize certain facial features such as the mouth (especially for "Happiness"), nose, eyes, and, in some cases, even the cheeks, chin, and forehead.
            
        \end{itemize}


\section{Discussion}


        \subsection{Can DL models, originally trained with standard datasets of facial expressions, yield proficient performance when applied to individuals with intellectual disabilities?(Q1)}

        The outcomes derived from the initial experiment indicate the impracticality of achieving proficient performance with DL models initially trained on conventional datasets of facial expressions when applied to individuals with intellectual disabilities. After training on an ensemble of five datasets for FER, the accuracy achieved by the different models remained high (approximately 90\%) despite changing the dataset used for evaluation to Google FE-Test, showcasing a proper generalization of the networks, and unlocking their use for unseen users and data. However, this adaptability did not extend to the MuDERI dataset, exclusively comprising individuals with intellectual disabilities, as evidenced by a significant drop in accuracy to below 50\% across all networks. This suggests that the facial expressions of individuals with intellectual disabilities exhibit unique characteristics that are not well captured by the standard datasets.

        Furthermore, the opposite case was also found to hold true: training DL models solely on a dataset of individuals with intellectual disabilities did not lead to good generalization performance on standard facial expression datasets. When models trained on the MuDERI dataset were evaluated on the Google FE-Test, the accuracy was also poor, approximately 40\%. This suggests that the facial expressions of individuals without intellectual disabilities also exhibit unique characteristics that are not well captured by the MuDERI dataset.

        The discernible dissimilarity in facial expressions between individuals with and without intellectual disabilities poses a considerable challenge for the twelve tested networks, hindering their efficacy in both directions. These results underscore the imperative for a more tailored training approach, specifically tuned for the facial expressions exhibited by individuals with intellectual disabilities. Additionally, research on the physiological and psychological factors that contribute to facial expressions in individuals with intellectual disabilities can provide insights for designing more effective training datasets and models.

        \bigskip
        
        \subsection{Can deep learning models, specifically trained on a dataset comprising individuals with intellectual disabilities, accurately predict the facial expressions of other individuals with intellectual disabilities? (Q2)}

        Looking at the figures in section \ref{sec:second_exp}, it is clear that it is not possible for any of the models to properly recognize the facial expression of users with intellectual disabilities unseen during the training phase, since the results in these cases are similar to that achieved when they are trained on users without disabilities: below 50\% accuracy for almost all networks, and below 30\% F1 score in all cases. This discrepancy suggests a divergence among users, somewhat analogous to the distinction between users from the MuDERI dataset and those from FER-DB5, without intellectual disabilities.

        However, the results also demonstrate that user-specific fine-tuning can significantly improve the recognition performance. When individuals to be tested are included in the training set, allowing the models to learn the unique facial expression patterns of each person, the accuracy increases considerably. This improvement is particularly evident when at least one clip per class is present for each individual during training, contributing to an approximate 10\% increase in the F1 score.

        Altogether, DL models trained solely on a dataset of individuals with intellectual disabilities struggle to generalize well to unseen individuals within the same population. This highlights the need for more personalized approaches, involving user-specific fine-tuning, to achieve accurate FER for individuals with intellectual disabilities. Such fine-tuning can be achieved by incorporating additional training data from the target individuals, enabling the models to learn their unique facial expression patterns.

        \bigskip
        
        \subsection{What disparities and commonalities exist in the facial expressions of individuals with and without intellectual disabilities, learned by DL models?(Q3)}
       
        To address this question, the third experiment employed a XAI technique to generate heat maps highlighting the facial regions that significantly influence the models' predictions. These heat maps (shown in Figure \ref{fig:heatmaps}) were analyzed across various model architectures, training datasets, and test datasets.

        Firstly, the dependence of heat maps on the training set is notable, overshadowing the influence of the test set, for which the explanations were computed. This effect can be observed on the heat maps for the Google FE-Test and MuDERI datasets, computed for the models trained on FER-DB5, and the logical inference is that the networks focus on the patterns they have learned in the training phase. Nonetheless, the disparate accuracy outcomes (approximately 90\% on Google FE-Test to less than 50\% on MuDERI) imply inadequacy of these learned patterns for facial expression recognition on MuDERI, which comprises users with intellectual disabilities.

        Secondly, heat maps exhibit variation across different networks, with more pronounced differences arising as architectures diverge. For instance, minimal disparity is observed between VGG16 and VGG19 models or among SilNet, SongNet, and WeiNet, whereas more substantial differences emerge between these two groups. Moreover, variability across models is more prominent when explaining the models trained on MuDERI, indicative of reduced consensus in finding solutions.
        
        Lastly, attention is drawn to the first and third columns for each class in the figure, allowing a comparison of learned regions between the FER-DB5 and MuDERI datasets. Heat maps computed for FER-DB5 highlight different regions of the face, which vary depending on the model and the expression: the mouth and nose are especially important across models and classes, but also the chin, forehead and cheeks for some cases, roughly aligning with human expectations. In contrast, MuDERI's heat maps appear more erratic, exhibiting greater variance across models and highlighting less intuitive regions. These less interpretable heat maps, coupled with increased variation between models, suggest MuDERI poses a more challenging dataset, demanding solutions that are both less generalizable and less intuitive.

    \subsection{Limitations of the study}

        In this study, we trained 12 distinct neural networks for facial expression recognition tasks on two datasets: FER-DB5, representing individuals without intellectual disabilities, and MuDERI, encompassing individuals with intellectual disabilities. We could not find any other public dataset including people with intellectual disabilities, however, we acknowledge several limitations associated with the MuDERI dataset:

        \begin{itemize}
        
        \item \textbf{Size}: While the dataset size was sufficient for training various models, increased data volume is essential for enhanced generalization and prediction stability.
            
        \item \textbf{Class imbalance}: The dataset exhibits an imbalance, with more samples for the "happiness" class than for others. Although this imbalance does not severely compromise predictions, it can impact performance, which is why the F1 score, considering both precision and recall, was utilized in this study to measure the models' performance.
            
        \item \textbf{Variety}: All dataset images originate from recordings of twelve participants, potentially leading to overfitting if not carefully addressed, as observed in the second experiment results (Section \ref{sec:second_exp}). To mitigate this, a more extensive range of users and recordings is preferable to diversify the dataset and reduce dependence on specific users, scenarios, or lighting conditions.

        \item \textbf{Quality}: In some instances, the camera perspective is suboptimal, either due to users not facing the camera or the camera being positioned too high.

        \end{itemize}

        A significant challenge faced in this study lies in the scarcity of data for the FER task concerning individuals with intellectual disabilities. Consequently, we utilized only one dataset as a representative sample for this population, contrasting with an ensemble of datasets for individuals without intellectual disabilities. Future endeavors should prioritize addressing this data scarcity by proposing new datasets that specifically include individuals with intellectual disabilities.

\section{Conclusion}
   
    This study explored the challenges associated with applying automatic FER to individuals with moderate to severe intellectual disabilities. Twelve different DL models were trained, exploring various dataset combinations and splits, including the use of the MuDERI dataset, comprised solely of users from this demographic. Additionally, explainability techniques were used to provide insights into the internal mechanisms of the models concerning users with and without intellectual disabilities.
    
    Results underscored the inadequacy of models trained on generic FER datasets that exclude individuals with intellectual disabilities. The findings emphasized the necessity for tailored training inclusive of this specific user group. Moreover, substantial variations were observed even within this demographic, emphasizing the importance of incorporating target users in the training set for optimal model performance. A detailed exploration using Explainable Artificial Intelligence (XAI) techniques uncovered significant differences in facial regions employed by the models for expression recognition when trained on users with and without intellectual disabilities. The patterns identified were more intricate for the former, featuring less intuitive regions.
    
    The main contributions of this work are:
   
     \begin{enumerate}
        \item We conducted a pioneer exploration on the application of existing automated FER systems to people with intellectual disabilities to study its direct use in systems such as social robots. We reported the performance results, with low values for unseen individuals with intellectual impairments in all cases. 
        \item We provided a global explanation in the form of heatmaps to identify the important face regions considered in decision making by the DL models. Results show the disparities on the important face regions for individuals with and without intellectual impairments. 
        \item Based on the results and the heatmaps, we concluded that extrapolating FER systems based on DL to individuals with intellectual impairments is not feasible. Despite training  with both people with and without intellectual impairments, the application to unseen individuals is not effective. Therefore, user-specific training should be done. 
    \end{enumerate}

   The results indicate that existing FER systems cannot be directly integrated in technology such as social robots, when used with individuals with intellectual disabilities as they may express themselves in a unique way. This variability could explain why the models do not generalize well across different users and why the critical facial regions differ from one user to another. To reassert these results, further studies with larger datasets are needed, however, research should also focus on the existing data scarcity for FER in individuals with intellectual disabilities.

\section*{Acknowledgments}
Grant PID2019-104829RA-I00 funded by MCIN/AEI/10.13039/ 501100011033, project EXPLainable Artificial INtelligence systems for health and well-beING (EXPLAINING). Grant PID2022-136779OB-C32 funded by MCIN/AEI/ 10.13039/501100011033 and by ERDF A way of making Europe, project Playful Experiences with Interactive Social Agents and Robots (PLEISAR): Social Learning and Intergenerational Communication. F. Xavier Gaya-Morey was supported by an FPU scholarship from the Ministry of European Funds, University and Culture of the Government of the Balearic Islands.

\bibliography{bibliography}
\bibliographystyle{IEEEtran}

\newpage

\section{Biography Section}
\begin{IEEEbiography}[{\includegraphics[width=1in,height=1.25in,clip,keepaspectratio]{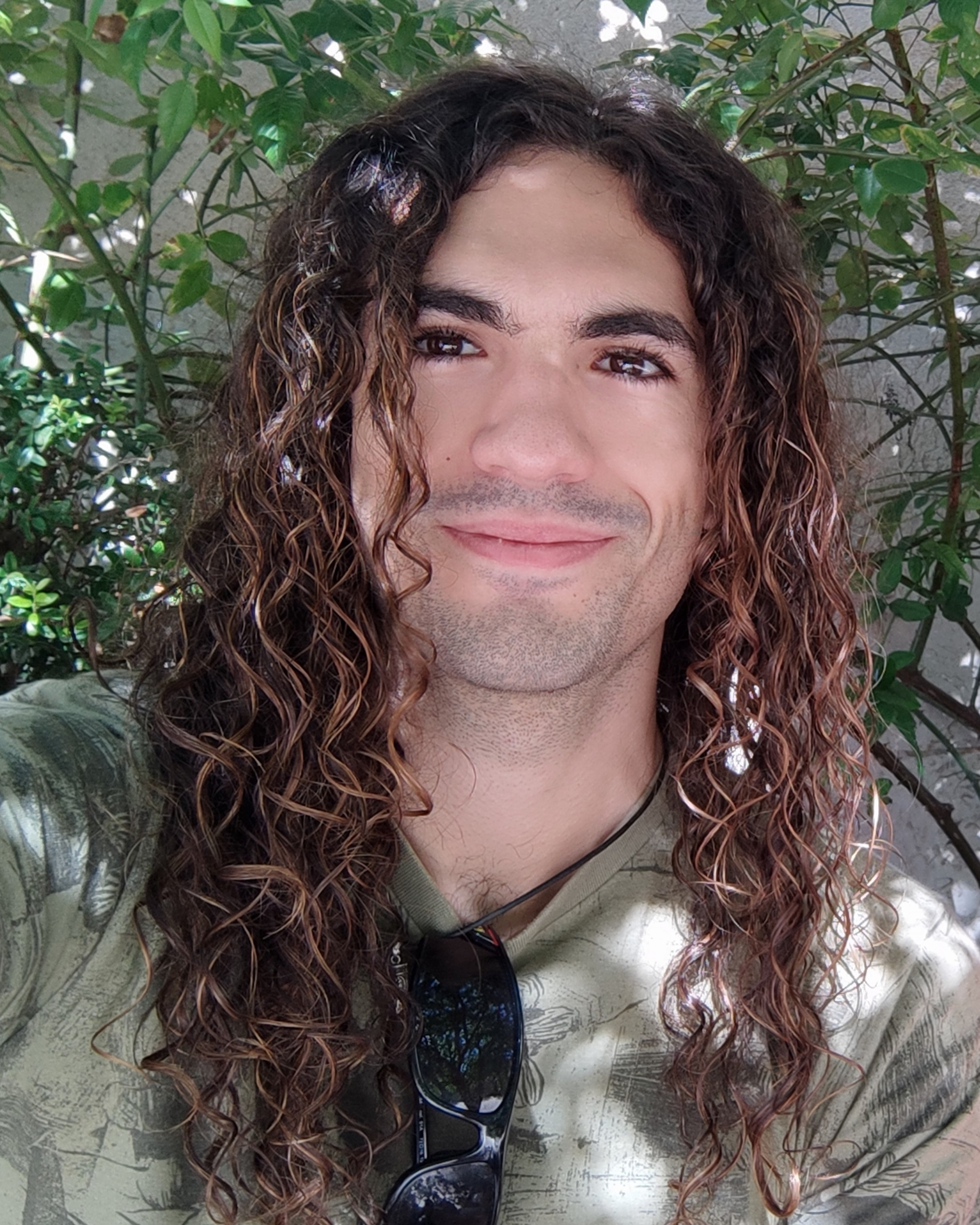}}]{F. Xavier Gaya-Morey}
F. Xavier Gaya-Morey is a Ph. D. candidate and professor at the Universitat de les Illes Balears. He holds a bachelor’s degree in computer engineering and a master’s degree in data science and computer vision. His research is centered on the areas of explainable artificial intelligence and computer vision, with special focus on their applications to improve the life quality of the older adults.
\end{IEEEbiography}

\begin{IEEEbiography}[{\includegraphics[width=1in,height=1.25in,clip,keepaspectratio]{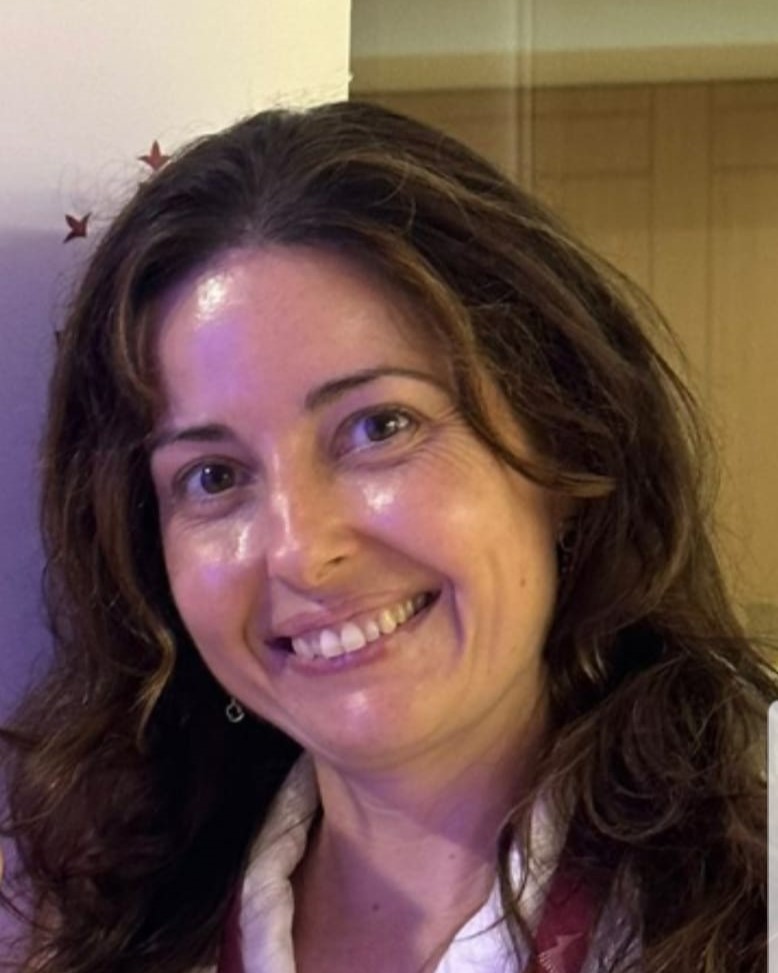}}]{Silvia Ramis}
Silvia Ramis, Ph. D. in Information and Communications Technologies from the UIB (since 2019). She has participated in several projects in the field of Computer Vision, Artificial
Intelligence, Explainable Artificial Intelligence and Human-Robot Interaction. Her research
experience focuses on artificial intelligence applied to human-robot interaction, especially in
face detection and facial expression recognition.
\end{IEEEbiography}

\begin{IEEEbiography}[{\includegraphics[width=1in,height=1.25in,clip,keepaspectratio]{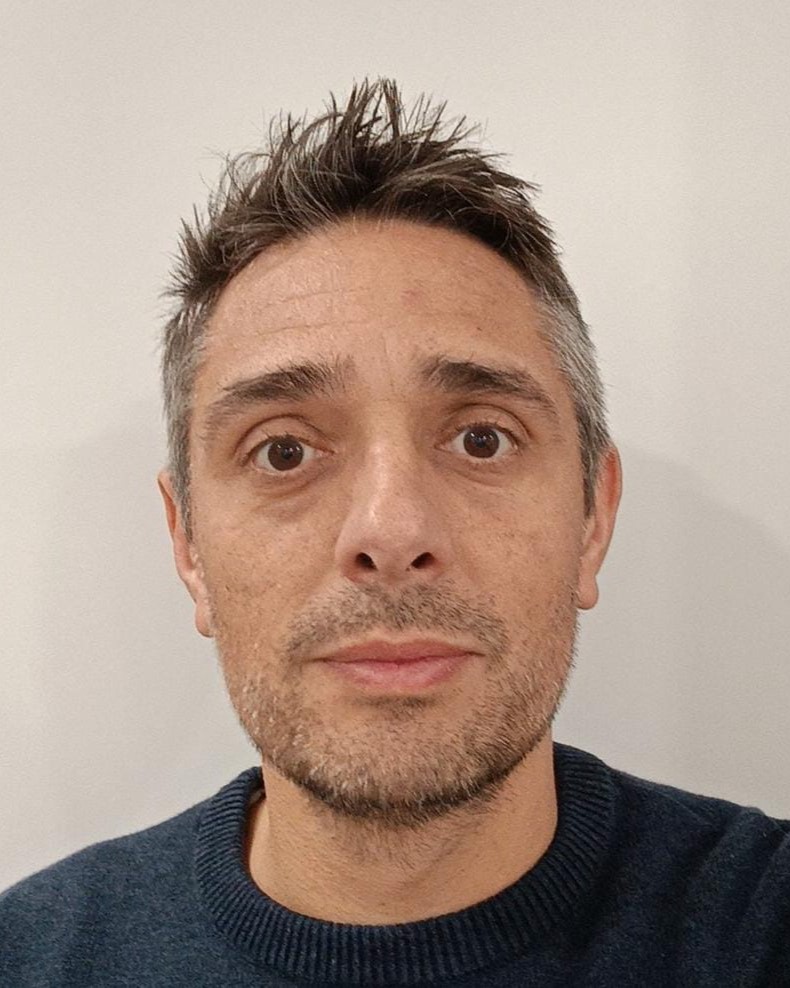}}]{Jose M. Buades-Rubio}
Jose M. Buades received his degree in Computer Science and his Ph. D. in Computer Science from the University of Balearic Islands. He is currently an Associate Professor at the University of the Balearic Islands. His research interests include computer graphics, computer vision and artificial intelligence.
\end{IEEEbiography}

\begin{IEEEbiography}[{\includegraphics[width=1in,height=1.25in,clip,keepaspectratio]{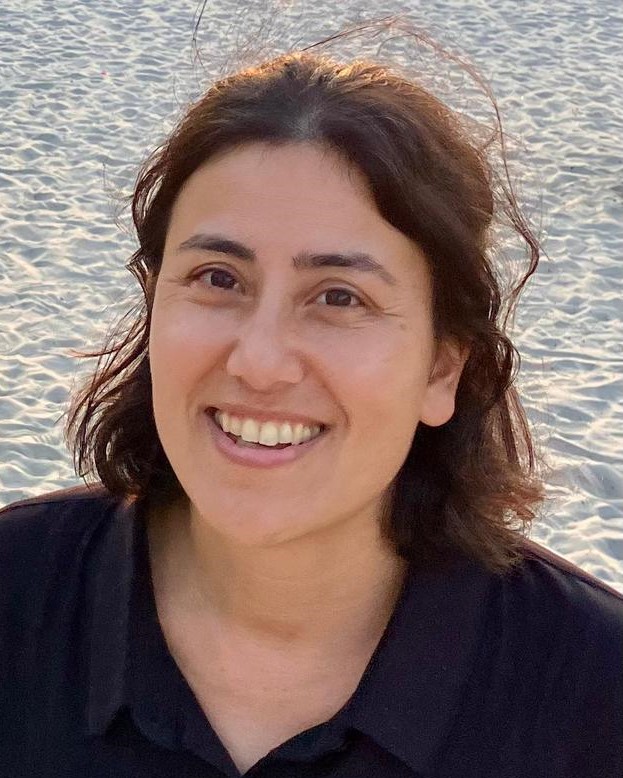}}]{Cristina Manresa-Yee}
Cristina Manresa-Yee received her degree in Computer Science and her Ph. D. in Computer Science from the University of Balearic Islands. She is currently an Associate Professor at the University of the Balearic Islands. Her research interests include human-computer interacQon, computer vision and explainable AI.
\end{IEEEbiography}

\vfill

\end{document}